\documentclass{article}
\usepackage{spconf,amsmath,graphicx,hyperref}
\usepackage{amssymb,booktabs,tikz}

\usetikzlibrary{positioning,fit,arrows.meta,calc}
\newcommand{\pmstd}[1]{{\,\tiny$\pm$#1}}
\newcommand{\first}[1]{\textbf{\boldmath#1}}

\newcommand{\mserow}[1]{{\scriptsize\color{black!62}#1}}

\newcommand{\wallsWaveNet}{21}
\newcommand{\wallsBSR}{20}
\newcommand{\aboveMixWaveNet}{19}
\newcommand{\aboveMixBSR}{16}

\providecommand{\wallsWaveNet}{21}
\providecommand{\wallsBSR}{20}
\providecommand{\aboveMixWaveNet}{19}
\providecommand{\aboveMixBSR}{16}

\title{Toward Interference-Agnostic RF Source Separation}

\name{\begin{tabular}{@{}c@{}}
Maxence Albert$^{\star}$ \quad Rapha\"el Duroselle$^{\star}$ \quad
Thibaud Ehret$^{\star}$ \\
Olivier Boeffard$^{\star}$ \quad Jean-Fran\c{c}ois Bonastre$^{\star}$
\end{tabular}}
\address{$^{\star}$AMIAD, Palaiseau, France}

\begin{document}
\ninept
\maketitle
\begin{abstract}
Separating a radio-frequency signal of interest from a co-channel
interference is mostly addressed with specialized separators, each trained
for one pair of signal and interference types and therefore relying on a
label that a heavily jammed receiver cannot identify. We show that this
reliance is not benign: applied to an interference it was not trained on,
such a separator usually fails to recover the signal, and in most of the
cases we measure it leaves the signal worse than no processing at all. We
then train a single separator over every pair, giving it no indication at
test time of which one it faces, on two architectures with no component in
common: the convolutional WaveNet of the ICASSP 2024 RF Challenge, and a
band-split transformer that we adapt from music source separation. Both
behave the same way. At the size of one specialized model, the unified
separator stays close to the whole models it replaces, keeping the bit
error rate close on almost every pair, better on one and worse on another. On an interference absent from every training set, it does
as well as the best specialized model, without anyone having to choose
which one to deploy.
\end{abstract}


\begin{figure}[t]
\centering
\includegraphics[width=\columnwidth]{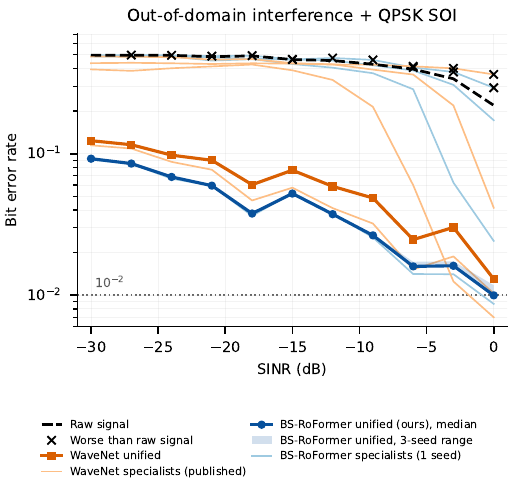}
\caption{Models applied to a mixture of a QPSK signal of interest and an
interference absent from every training set. The proposed unified band-split
transformer reaches out-of-domain performance as good as the best specialized
model, without having to choose which one to deploy. A poor choice degrades the
signal instead: three of the eight specialists are worse than the raw signal at
one or more points, marked $\times$.}
\label{fig:ood}
\end{figure}

\section{Introduction}
\label{sec:intro}

Radio frequency (RF) receivers increasingly operate in bands shared with uncontrolled RF emitters. When a third-party signal occupies the same band as the signal of interest (SOI) and the receiver has a single antenna, the two arrive summed on a single channel, with no spatial diversity to separate them. Classical approaches such as matched filtering~\cite{turin1960matched},
equalization~\cite{qureshi1985equalization} and Wiener
filtering~\cite{wiener1949} rely on linear or second-order processing, which do not
capture the structure of a digital interference. Learned separators exploit
that structure instead, and have been applied to single-channel RF separation
in several forms: score-based generative models~\cite{jayashankar2023score},
architectures built around the OFDM structure of the signal of
interest~\cite{lee2023ofdmarch}, and, most often, networks transposed from
audio and image source separation. What this literature long lacked is a common
measurement point.

The ICASSP 2024 RF Challenge~\cite{lancho2024rfchallenge} provides one. It introduces a dataset of two synthetic SOIs, a root-raised-cosine QPSK waveform and an OFDM (orthogonal frequency-division multiplexing) waveform, each mixed with four recorded interferences, which gives eight (SOI, interference) pairs, and it scores a system on the bit error rate (BER) of the demodulated SOI and on the mean square error (MSE, in dB) of the estimate. Its reference models, a WaveNet and a U-Net~\cite{lancho2024rfchallenge}, and its leading
entries~\cite{tian2024wavenet,damara2024selfattention} share one design: one
network per pair, trained on that pair alone. 

That design is reasonable in a competition, where the pair is given. In deployment the interference is whatever occupies the band, and selecting the right model requires knowing its type, which the receiver cannot verify. The cost of that assumption has not, to our knowledge, fully been measured on this problem.

Separators trained on several tasks at once are standard in
audio~\cite{kavalerov2019universal,tzinis2020improving,saijo2025tuss}, and
recent RF work has begun to address an uncertain interference type: unified
models have been trained over the four benchmark interferences with a fixed
QPSK SOI~\cite{rftransformer}, and detect-then-separate pipelines have been
compared with unified models on two interferences at a time with PSK signals of
interest, using an oracle interference classifier~\cite{rodrigez2026uncertainty}.
Two questions central to deployment remain open: what removing the label costs
when one separator must also cover signals of interest from different waveform
families, whose type may itself not be identifiable under strong interference,
and how badly a specialist fails when the interference is not the one it was
trained on.

This paper makes two contributions. First, on the eight pairs of the
challenge, QPSK and OFDM, we quantify pair by pair what a single
\emph{unified} separator loses against a family of specialists, and what it
avoids: a specialist deployed on the wrong interference rarely reaches
a BER of $10^{-2}$, a cost that evaluations restricted to correctly matched
specialists do not show (Secs.~\ref{ssec:unif}--\ref{ssec:identity},
Figs.~\ref{fig:ood} and~\ref{fig:identity}). We establish this on two
architectures with no component in common, so that the result is a property of
the training scheme rather than of a network. Second, one of the two is new to
the task: we adapt the band-split transformer of music source
separation~\cite{luo2023bsrnn,lu2024bsroformer} to complex RF signals
(Sec.~\ref{ssec:architecture}), where, trained per pair, it is on par with the
reference WaveNet on seven pairs and surpasses it by at least $8.6$\,dB on
QPSK+CS5G1, at eleven times the compute.\footnote{Code, frozen configurations, checkpoints and the evaluation
harness will be released with the final version of the paper.}

\begin{figure}[tb]
\centering
\begin{tikzpicture}[
  font=\scriptsize,
  scale=0.95, every node/.style={transform shape},
  bx/.style={draw=black!55, line width=0.4pt, rounded corners=1.5pt,
             fill=black!4, minimum height=5.5mm, minimum width=9mm,
             inner sep=1.5pt, align=center},
  op/.style={draw=black!55, line width=0.4pt, circle, fill=white,
             inner sep=0pt, minimum size=4mm},
  sub/.style={draw=black!35, line width=0.4pt, rounded corners=1.5pt,
              fill=white, minimum height=5.5mm, inner sep=2pt, align=center},
  ar/.style={-{Stealth[length=1.7mm]}, line width=0.45pt, black!70},
  lbl/.style={font=\tiny, inner sep=1pt, black!75},
  tag/.style={font=\tiny, inner sep=0pt, black!75, fill=white},
]
\node[lbl] (x) {$y\in\mathbb{C}^{L}$};
\node[bx, right=2mm of x] (stft) {STFT\\[-1pt]};
\node[bx, right=2mm of stft] (bs) {Band split\\[-1pt]};
\node[bx, right=2mm of bs, minimum width=14mm, minimum height=8mm]
  (blocks) {$N{=}12$ dual-\\[-1pt]path blocks};
\node[bx, right=2mm of blocks] (mask) {Mask\\[-1pt]head};
\node[op, right=5mm of mask] (mul) {$\odot$};
\node[bx, right=5mm of mul] (istft) {iSTFT};
\node[lbl, right=2mm of istft] (shat) {$\hat{s}$};
\draw[ar] (x) -- (stft);
\draw[ar] (stft) -- (bs);
\draw[ar] (bs) -- (blocks);
\draw[ar] (blocks) -- (mask);
\draw[ar] (mask) -- node[tag, above=0.4mm]{$M$} (mul);
\draw[ar] (mul) -- node[tag, above=0.4mm]{$\hat{S}$} (istft);
\draw[ar] (istft) -- (shat);
\coordinate (rlow) at ($(stft.south)+(0,-6mm)$);
\draw[ar] (stft.south) -- (rlow) -| node[tag, pos=0.35, below=0.4mm]{$Y$}
  (mul.south);
\node[sub, below=10mm of blocks, xshift=-9mm] (tattn)
  {Time attn\\[-1pt]\tiny RoPE};
\node[sub, right=2.5mm of tattn] (battn) {Band attn};
\node[sub, right=2.5mm of battn] (ffn) {FFN};
\node[draw=black!35, line width=0.4pt, dashed, rounded corners=2pt,
      fit=(tattn)(battn)(ffn), inner sep=2mm] (zoom) {};
\draw[ar] (tattn) -- (battn);
\draw[ar] (battn) -- (ffn);
\node[lbl, above right=0.2mm and 0mm of zoom.north west] {one block};
\draw[black!35, line width=0.35pt, dashed,
      preaction={draw, white, line width=1.6pt}]
  (blocks.south) -- (zoom.north);
\end{tikzpicture}
\caption{The proposed band-split separator. The mixture $y$ is analyzed by a
short-time Fourier transform (STFT), its bins are split into $K=16$ uniform
bands and the result is processed by $N=12$ blocks that alternate attention
along time with attention across bands. The head produces a complex ratio mask
$M$ of unbounded magnitude, applied bin by bin to the spectrogram $Y$ of the
mixture; the inverse transform returns the estimated signal of interest
$\hat{s}$. Sec.~\ref{ssec:architecture} details each stage.}
\label{fig:arch}
\end{figure}
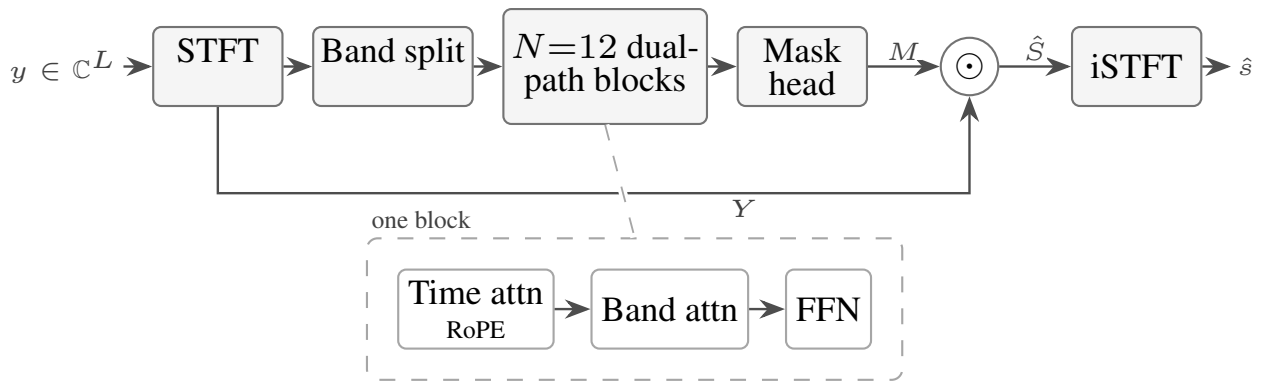

\section{Interference-agnostic source separation}
\label{sec:model}

A mixture is the sum of two complex RF signals of length $L$, \begin{equation}
y[n] = s[n] + \kappa\, b[n], \qquad n = 0,\dots,L-1,
\label{eq:mix}
\end{equation}
where $s$ is the SOI, $b$ the interference and $\kappa>0$ a per-example scale
that sets the signal-to-interference-plus-noise ratio (SINR), 
$10\log_{10}\big(\|s\|_2^2/\|\kappa b\|_2^2\big)$. A separator estimates $s$
from $y$ alone. The lower the SINR, the more the interference dominates the
samples the demodulator sees and the harder the recovery, so performance is a
curve over SINR rather than a single number.

Existing separators train one network per pair. We train one network over all pairs instead, and give it no indication of which one it faces. Because this is a statement about training
and not about a network, we evaluate it on two architectures that share nothing
but their input: the fully convolutional WaveNet used as a reference by the
challenge, and a band-split transformer that we adapt here from music source
separation.

\subsection{Two architectures}
\label{ssec:architecture}

Single-channel RF separation has largely borrowed its architectures from audio,
where the same object is processed: a one-dimensional sequence with two
channels, left and right for a stereo waveform, in-phase and quadrature for a
complex RF signal.

\noindent\textbf{WaveNet.} Our first architecture is the challenge reference~\cite{lancho2024rfchallenge}.
It is fully convolutional~\cite{oord2016wavenet}, a stack of dilated convolutions applied to the
time samples with no attention and no explicit frequency representation.
It is the most accurate published system with open weights and code. This allowed us to run the models on the public TestSet1.

\noindent\textbf{Band-split transformer.} Our second architecture is taken from
another family. Band-split models cut the spectrum into bands, keep one
token per band and per analysis frame, and alternate attention along time
inside a band with attention across bands at a given frame. They lead music source
separation~\cite{luo2023bsrnn,lu2024bsroformer,wang2023melband}. We adapt the
band-split RoPE transformer (BS-RoFormer) of~\cite{lu2024bsroformer} to complex RF signals, Fig.~\ref{fig:arch} shows the architecture.

The mixture is analyzed by a short-time Fourier transform (STFT) with a Hann
window of $N_{\mathrm{fft}}=64$ samples, that is $3.2\,\mu$s at the $20$\,MS/s
of the challenge, and a hop of $16$ samples, or $0.8\,\mu$s. The FFT is taken on
the window length, and the input being complex the
spectrum is two-sided: $64$ bins of $312.5$\,kHz, which is also the subcarrier
spacing of the OFDM SOI. The bins are cut into $K=16$ uniform bands of four,
each band-frame token is projected into dimension $D$ by weights of its own,
and the resulting $K\times T\times D$ tensor goes through $N$ dual-path blocks \cite{chen2020dptnet}.
Each block is three residual sublayers: attention along the $T$ frames inside a
band, with rotary positional encoding~\cite{su2024roformer}; attention across the
$K$ bands at a fixed frame, with no positional encoding, the band order being
already carried by the per-band projections; and a position-wise feed-forward
network. A linear head returns a complex ratio
mask~\cite{williamson2016crm} $M\in\mathbb{C}^{N_{\mathrm{fft}}\times T}$, and
the estimate is $\hat{S} = M \odot Y$, $\hat{s} = \mathrm{iSTFT}(\hat{S})$.

Two choices depart from the audio design. The bands are uniform, where music
separators use non-uniform layouts, narrow at low frequency and wide at
high~\cite{luo2023bsrnn,lu2024bsroformer} or mel-spaced~\cite{wang2023melband}. 
And the magnitude of $M$ is left unbounded, where
audio band-split models bound it.
RF interference can be destructive, so $|Y|<|S|$ on a fraction of the bins, and a
bounded mask cannot restore their magnitude. Both were fixed by a $50$k-step ablation on the two EMI pairs.

\subsection{Specialist, unified and conditioned models}
\label{ssec:systems}

We call \emph{specialist} a model trained on a single (SOI, interference)
pair, the design of the reference models and of the leading entries of the
challenge, and \emph{unified} a model trained on the union of the eight pairs,
evaluated on each of them.

Part of the gap between a unified model and a specialist may come from attributes of the mixture that the network must infer rather than receive. To measure how much a correct label is worth, we equip the unified model with optional conditioning: two learnable embedding tables, one for the SOI type and one for the interference type, summed and added to the input of the block stack.

\section{Experiments}
\label{sec:exp}
\begin{table*}[t]
\centering
\caption{Source separation performance on TestSet1. \textsc{ber}$\downarrow$ is
the BER score of Sec.~\ref{ssec:setup}, the SINR in dB at which the BER curve
crosses $10^{-2}$; \texttt{--} a curve that never reaches it; a $-30.0$ entry
marks the left edge of the grid, so the true BER score is lower.
\mserow{\textsc{mse}$\downarrow$} is the official MSE score in dB. Spec.\ is one
model per pair, Unified one model for the eight, mid and small the same at
$D{=}256$, $N{=}8$ and $D{=}144$, $N{=}6$; ours use $D{=}384$, $N{=}12$ unless
stated. Unified and Unified cond.: mean and standard deviation over three
seeds; every other column is a single run. Unified cond.\ is given the
interference and SOI labels, and is therefore excluded from the comparison.
\first{Bold}: best of the six other columns, separately for \textsc{ber} and
\textsc{mse}.}
\label{tab:main}
\vspace{6pt}
\footnotesize\setlength{\tabcolsep}{1.2pt}
\resizebox{\linewidth}{!}{
\begin{tabular}{l cc || cc | cc@{\hspace{3pt}} cc@{\hspace{3pt}} cc@{\hspace{3pt}} cc@{\hspace{3pt}} cc}
\toprule
 & \multicolumn{2}{c||}{Published~\cite{lancho2024rfchallenge}} & \multicolumn{12}{c}{This paper} \\
\cmidrule(lr){2-3}\cmidrule(l){4-15}
 & \multicolumn{2}{c||}{WaveNet} & \multicolumn{2}{c|}{WaveNet} & \multicolumn{10}{c}{Band-split transformer} \\
\cmidrule(lr){4-5}\cmidrule(l){6-15}
 & \multicolumn{2}{c||}{Spec.} & \multicolumn{2}{c|}{Unified} & \multicolumn{2}{c}{Spec.} & \multicolumn{2}{c}{Unified} & \multicolumn{2}{c}{Unified mid} & \multicolumn{2}{c}{Unified small} & \multicolumn{2}{c}{Unified cond.} \\
\cmidrule(lr){2-3}\cmidrule(lr){4-5}\cmidrule(lr){6-7}\cmidrule(lr){8-9}\cmidrule(lr){10-11}\cmidrule(lr){12-13}\cmidrule(l){14-15}
Pair & \textsc{ber}$\downarrow$ & \mserow{\textsc{mse}$\downarrow$} & \textsc{ber}$\downarrow$ & \mserow{\textsc{mse}$\downarrow$} & \textsc{ber}$\downarrow$ & \mserow{\textsc{mse}$\downarrow$} & \textsc{ber}$\downarrow$ & \mserow{\textsc{mse}$\downarrow$} & \textsc{ber}$\downarrow$ & \mserow{\textsc{mse}$\downarrow$} & \textsc{ber}$\downarrow$ & \mserow{\textsc{mse}$\downarrow$} & \textsc{ber}$\downarrow$ & \mserow{\textsc{mse}$\downarrow$} \\
\midrule
QPSK+CS2 & $-18.01$ & \mserow{$-24.93$} & \first{$-19.66$} & \mserow{\first{$-26.37$}} & $-18.06$ & \mserow{$-21.49$} & $-19.63$\pmstd{0.01} & \mserow{$-22.17$\pmstd{0.10}} & $-19.50$ & \mserow{$-20.04$} & $-19.23$ & \mserow{$-16.67$} & $-17.99$\pmstd{0.03} & \mserow{$-22.07$\pmstd{0.12}} \\
QPSK+CS3 & \first{$-1.83$} & \mserow{\first{$-4.59$}} & $-1.53$ & \mserow{$-4.31$} & $-1.23$ & \mserow{$-3.69$} & $-0.77$\pmstd{0.02} & \mserow{$-2.55$\pmstd{0.02}} & $-0.61$ & \mserow{$-2.21$} & $-0.56$ & \mserow{$-2.14$} & $-1.57$\pmstd{0.20} & \mserow{$-1.80$\pmstd{1.02}} \\
QPSK+CS5G1 & $-21.43$ & \mserow{$-29.79$} & $-16.08$ & \mserow{$-18.76$} & \first{$-30.00$} & \mserow{\first{$-33.91$}} & $-24.58$\pmstd{1.66} & \mserow{$-23.04$\pmstd{2.02}} & $-26.68$ & \mserow{$-23.11$} & $-6.42$ & \mserow{$-8.13$} & $-29.31$\pmstd{0.61} & \mserow{$-29.91$\pmstd{0.23}} \\
QPSK+EMI & $-25.81$ & \mserow{\first{$-33.31$}} & $-25.15$ & \mserow{$-27.31$} & \first{$-26.00$} & \mserow{$-30.32$} & $-25.17$\pmstd{0.02} & \mserow{$-26.35$\pmstd{0.17}} & $-24.37$ & \mserow{$-23.49$} & $-22.79$ & \mserow{$-19.13$} & $-25.61$\pmstd{0.02} & \mserow{$-29.01$\pmstd{0.26}} \\
OFDM+CS2 & \first{$-8.84$} & \mserow{$-6.59$} & $-8.63$ & \mserow{$-6.50$} & $-8.43$ & \mserow{$-5.29$} & $-8.19$\pmstd{0.02} & \mserow{\first{$-6.73$}\pmstd{0.04}} & $-7.84$ & \mserow{$-6.15$} & $-6.77$ & \mserow{$-5.14$} & $-8.38$\pmstd{0.03} & \mserow{$-8.30$\pmstd{0.13}} \\
OFDM+CS3 & -- & \mserow{\first{$-2.32$}} & -- & \mserow{$-1.81$} & -- & \mserow{$-0.06$} & -- & \mserow{$-0.45$\pmstd{0.03}} & -- & \mserow{$-0.32$} & -- & \mserow{$-0.25$} & -- & \mserow{$-0.38$\pmstd{0.12}} \\
OFDM+CS5G1 & $-10.51$ & \mserow{$-10.85$} & -- & \mserow{$-9.29$} & \first{$-12.25$} & \mserow{\first{$-11.99$}} & -- & \mserow{$-8.89$\pmstd{0.03}} & -- & \mserow{$-8.40$} & -- & \mserow{$-7.76$} & -- & \mserow{$-9.65$\pmstd{0.09}} \\
OFDM+EMI & $-16.35$ & \mserow{$-14.77$} & $-15.06$ & \mserow{$-11.58$} & \first{$-16.72$} & \mserow{\first{$-18.03$}} & $-14.53$\pmstd{0.02} & \mserow{$-13.07$\pmstd{0.07}} & $-13.96$ & \mserow{$-12.00$} & $-13.29$ & \mserow{$-10.81$} & $-15.20$\pmstd{0.03} & \mserow{$-14.55$\pmstd{0.05}} \\
\midrule
Pairs with a BER score & \multicolumn{2}{c||}{$7/8$} & \multicolumn{2}{c|}{$6/8$} & \multicolumn{2}{c}{$7/8$} & \multicolumn{2}{c}{$6/8$} & \multicolumn{2}{c}{$6/8$} & \multicolumn{2}{c}{$6/8$} & \multicolumn{2}{c}{$6/8$} \\
Params (M) / \mserow{MFLOPs per sample} & \multicolumn{2}{c||}{$8{\times}3.96$ / \mserow{$7.90$}} & \multicolumn{2}{c|}{$3.96$ / \mserow{$7.90$}} & \multicolumn{2}{c}{$8{\times}31.01$ / \mserow{$87.09$}} & \multicolumn{2}{c}{$31.01$ / \mserow{$87.09$}} & \multicolumn{2}{c}{$9.62$ / \mserow{$27.14$}} & \multicolumn{2}{c}{$2.40$ / \mserow{$7.16$}} & \multicolumn{2}{c}{$31.01$ / \mserow{$87.09$}} \\
\bottomrule
\end{tabular}
}
\end{table*}
\subsection{Setup}
\label{ssec:setup}

\noindent\textbf{Data.} We evaluate every system with the protocol of the ICASSP 2024 RF challenge~\cite{lancho2024rfchallenge}. We use the public training set and the public test set, TestSet1. There also exists a private test set on which participants in the RF challenge were evaluated \cite{tian2024wavenet,damara2024selfattention}. The four interferences, CommSignal2, CommSignal3, CommSignal5G1 and EMISignal1, are written CS2, CS3, CS5G1 and EMI. Our validation split is drawn from the training set and is disjoint from TestSet1. Each pair is evaluated at eleven SINR points from $-30$ to $0$\,dB in $3$\,dB steps, with one hundred signals of $L=40\,960$ complex samples per point.

\noindent\textbf{Unseen interference.} To test the models on an interference absent from training, we use DroneDetect~\cite{swinney2021dronedetect}, OFDM IQ recordings of seven consumer drones at $2.4$\,GHz in three flight modes. We low-pass filter and decimate them from $60$\,MHz to the $20$\,MHz sample rate of the challenge (occupied band about $9.4$\,MHz at $-10$\,dB) and power-calibrate them like the official interferences. This test uses the QPSK SOI only.

\noindent\textbf{Models.} Table~\ref{tab:main} reports seven models. The WaveNet
specialists are the eight trained models released with the
challenge~\cite{lancho2024rfchallenge}, evaluated as published. The unified
WaveNet does not exist in the challenge and we trained it ourselves as described below. For the band-split
transformer we trained the eight specialists and three unified models of
decreasing size, so that the effect of unification can be separated from the
effect of capacity. The seventh is the conditioned variant of Sec.~\ref{ssec:systems}, at the largest size.

\noindent\textbf{Training.} Our large models use $D=384$, $N=12$ and $8$ attention heads, the
configuration of the original BS-RoFormer~\cite{lu2024bsroformer}; with $16$
bands instead of $62$, we have $31.0$\,M parameters, close to the eight
WaveNet specialists together ($8\times3.96=31.7$\,M). The loss is the sum, with equal weights, of the time-domain MSE between $\hat{s}$ and $s$ and a multi-resolution STFT loss~\cite{yamamoto2020parallel,guso2022loss}. The latter is computed at three FFT sizes, $64$ (the analysis STFT of Sec.~\ref{ssec:architecture}), $256$ and
$1024$; at each size it adds a spectral convergence, a log-magnitude and a
complex-spectrum term of weight $1$, and a phase term of weight $0.5$. We use
AdamW~\cite{loshchilov2019adamw} ($\beta=(0.9,0.95)$, weight decay $10^{-4}$,
gradient-norm clipping at $0.5$, \texttt{bf16}), batches of $256$ windows of
$5120$ samples, a peak learning rate of $3\cdot10^{-4}$ ($6\cdot10^{-4}$ for
the small model), a linear warm-up over the first $5\%$ of steps and a cosine
decay to $10^{-5}$. The SINR of each training window is drawn uniformly in
$[-33,3]$\,dB. Convergence is reached before $200$k steps for a specialist and
before $800$k for a unified model. The WaveNet keeps its published scheme: Adam~\cite{kingma2015adam}, no weight decay, time-domain MSE only, batches of
$32$ signals of $40\,960$ samples, \texttt{fp32}, for the same $800$k steps.

\noindent\textbf{Metrics.} We use the two scores of the
challenge~\cite{lancho2024rfchallenge}, per pair rather than summed over the
eight pairs. The \emph{BER score} is the lowest SINR at which the BER is at
most $10^{-2}$. We interpolate log-linearly between grid points,
to separate models that the $3$\,dB grid ties. A curve that stays above $10^{-2}$ over the whole grid has no BER score, written \texttt{--}.  The \emph{MSE score} is the MSE in dB averaged over the eleven SINR points. Lower is better for both.

\noindent\textbf{Inference.} Our models process each $40\,960$-sample test signal by overlap-add over the $5120$-sample training window, with $25\%$ overlap. The WaveNet processes the signal in one pass.

\subsection{Evaluation of specialist models}
\label{ssec:perf}
Trained per pair, the band-split transformer is competitive with the WaveNet,
with a large gain on QPSK+CS5G1 only, where it requires a larger model. We
compare the specialist models of the two architectures in
Table~\ref{tab:main}. The band-split transformer outperforms WaveNet on five
pairs of modulation and interference and is worse on two by at most
$0.60$\,dB. Neither architecture reaches a BER of $10^{-2}$ on OFDM+CS3. The
largest gain of the band-split transformer is observed on QPSK+CS5G1. An
ablation study shows that scaling up the model is necessary to obtain this
gain, and that this behavior is specific to this pair.

\subsection{Comparison of unified and specialist models}
\label{ssec:unif}

For both architectures, the unified model remains close to the eight
specialists it replaces and falls short on the same two pairs. Both keep the
$10^{-2}$ operating point on six pairs and lose it on OFDM+CS5G1; both lose
about $5$\,dB on QPSK+CS5G1, $5.35$\,dB for the WaveNet and at least $5.42$\,dB
for the transformer; and on the five remaining pairs both stay within
$2.2$\,dB of their own specialists, in either direction. Our unified
transformer is in fact better than the corresponding specialist on QPSK+CS2,
by $1.57$\,dB, as is the unified WaveNet, by $1.65$\,dB.

This deficit is a property of the pair, not of the network or of its capacity.
The unified WaveNet, with its own optimizer and objective, and our three
unified transformers, from $31.01$ down to $2.40$\,M parameters, reach a BER of
$10^{-2}$ on the same six pairs and miss the same two. Neither the architecture
nor a thirteenfold change of capacity moves that set.

The unified transformer retains the advantage of its architecture on
QPSK+CS5G1: it leads the unified WaveNet by $8.50$\,dB, and by $6.7$\,dB even
on its worst seed. On the five other pairs both reach a BER of
$10^{-2}$, the unified WaveNet
stays within $0.76$\,dB of it and is ahead on four, at one eleventh of the
compute. At matched compute, the WaveNet is better than our small unified
transformer on all six pairs, by $0.43$ to $9.66$\,dB.

Knowing the true pair by conditioning has a limited effect on most pairs and does not recover
OFDM+CS5G1. Given the true pair label, the unified conditioned transformer gains $0.19$ to
$0.80$\,dB on four pairs, but gains $4.7$\,dB on QPSK+CS5G1 and loses
$1.64$\,dB on QPSK+CS2, two differences that exceed the spread of the three
seeds. Sec.~\ref{ssec:identity} measures the cost of a wrong label.

On OFDM+CS5G1, the missing operating point is carried by four subcarriers out
of $56$, in both architectures. We score separately the four active
subcarriers closest to the center of the band, which hold $7\%$ of the bits,
for both specialists, the unified WaveNet, and the conditioned and mid-size
unified transformers. Over the SINR points above $-15$\,dB, these four
subcarriers carry $72$ to $77\%$ of the errors of every model, specialists
included, with a BER $20$ to $67$ times that of the other $52$. Unification
amplifies this weakness rather than creating a new one: from $-12$\,dB,
$67$ to $83\%$ of the errors that a unified transformer adds to its specialist
fall on these four subcarriers, and $58$ to $69\%$ for the WaveNet. On the other
$52$ subcarriers alone, every unified model is below $10^{-2}$ from $-12$\,dB.
Why joint training concentrates its penalty there remains open.

\subsection{Unification robustness analysis}
\label{ssec:identity}

Outside the pair it was trained on, a specialist rarely recovers the signal
and often degrades it, whereas the unified models never degrade it on an
unseen interference on the QPSK SOI. Table~\ref{tab:main} says nothing about this deployment
case, which we address with two experiments: specialists applied to another
interference of the benchmark, which models a receiver that identifies the SOI
but misidentifies the interference, and all models applied to drone traffic,
an interference absent from every training set.

\begin{figure}[tb]
\centering
\includegraphics[width=\columnwidth]{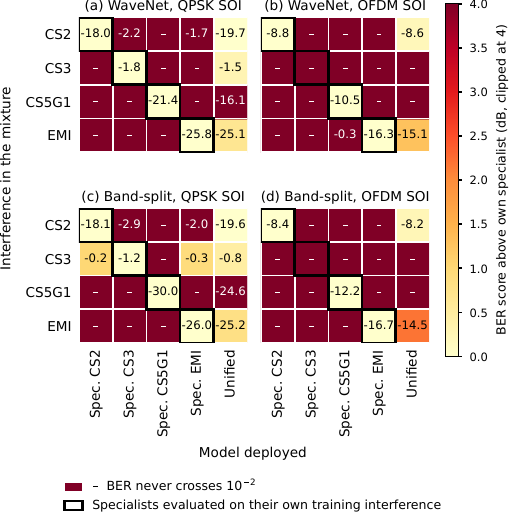}
\caption{Evaluation of the specialists and unified model on different interference. Rows: interference actually mixed with the SOI. Columns: model deployed, the four specialists of the SOI and, last column, the unified model. Cell: BER score in dB, \texttt{--} when the curve never reaches $10^{-2}$. Color: BER score minus that of the row's own specialist, in dB, clipped at $4$. Top: WaveNet. Bottom: band-split transformer. Left: QPSK SOI. Right: OFDM SOI.
Only the interference is mismatched: the SOI is assumed correctly identified.}
\label{fig:identity}
\end{figure}

\noindent\textbf{Interference recognition error (Fig.~\ref{fig:identity}).}
A receiver that picks the wrong specialist does not merely lose a few
decibels: in most cases, the specialist never reaches a BER of $10^{-2}$.
Each specialist is evaluated on the three other interferences with the same
SOI, which gives $24$ mismatched cells per architecture. For the WaveNet,
\wallsWaveNet{} of them never reach $10^{-2}$ and \aboveMixWaveNet{} are worse
than the raw signal at one SINR point or more. For the transformer, the counts
are \wallsBSR{} and \aboveMixBSR{}. The few mismatches that still reach
$10^{-2}$ involve CS2, CS3 and EMI on the QPSK SOI, and only the CS5G1 WaveNet
specialist applied to EMI on the OFDM SOI, no specialist other than its own
reaches $10^{-2}$ on CS5G1. This is the cost of a detection error in a
detect-then-separate pipeline~\cite{rodrigez2026uncertainty}, which that work
evaluates with an oracle interference classifier.

\noindent\textbf{Unseen interference (Fig.~\ref{fig:ood}).} On out-of-domain interference,
the unified models perform as well as the best specialist without having to
choose which one to deploy, while a poor choice of specialist degrades the
signal. Fig.~\ref{fig:ood} shows the BER against SINR on the drone mixtures
for the raw signal, the four QPSK specialists of each architecture and the two
unified models. At $0$\,dB, the raw signal is at a BER of $2.20\times10^{-1}$. The
unified transformer brings it to between $9.8\times10^{-3}$ and
$1.18\times10^{-2}$ over three seeds, the unified WaveNet to
$1.30\times10^{-2}$, and neither is worse than the raw signal at any of the
eleven points. Three of the eight specialists are: at $0$\,dB, the CS5G1
specialist returns $2.92\times10^{-1}$ in our architecture and
$3.63\times10^{-1}$ in the WaveNet.

\section{Limitations}
\label{sec:limits}
The protocol is narrow: two SOI, both synthetic and from the
same generator, four recorded interferences, and one out-of-domain interference
on one SOI. Nothing in it shows that a separator trained this way generalizes
to an arbitrary interference, a usable receiver would require more modulations and more
classes of emitter. The large unified transformer and its conditioned variant
have three seeds: the standard deviation of their BER score is at most
$0.03$\,dB, except on QPSK+CS5G1 ($1.66$ and $0.61$\,dB) and on QPSK+CS3 for
the conditioned variant ($0.20$\,dB). Every other model rests on a single run.

\section{Conclusion}
We have shown that a single separator, trained over the eight (signal,
interference) pairs of the ICASSP 2024 RF Challenge~\cite{lancho2024rfchallenge} and given no indication of
the pair it faces, can replace a bank of specialists with a limited BER cost on most benchmark pairs. This cost
is the same for two architectures with no component in common, the
convolutional WaveNet of the challenge and a band-split transformer: a
specialist reaches a BER of $10^{-2}$ on seven of the eight pairs, the unified
model on six, within $2.2$\,dB of the specialist on five of them. In return,
the receiver no longer has to identify the signal or the interference. A
specialist given the wrong interference never reaches $10^{-2}$ in
\wallsWaveNet{} and \wallsBSR{} of the $24$ cases, and on an unseen
interference only specialists degrade the QPSK signal.

We have also shown that the band-split transformer, adapted from music source
separation to RF signals, is a competitive architecture for this task.
Trained per pair, it is on par with the WaveNet on every pair and is the only
model to keep the BER below $10^{-2}$ over the whole grid on QPSK+CS5G1, at
eleven times its compute.

Whether these results hold beyond two synthetic signals of interest and one
unseen interference remains open. The objective is a universal receiver that replaces a
bank of detectors and specialized separators by one network and holds in a band
it has never seen.
\vspace{1ex}

\noindent\textbf{Acknowledgments.} This project was provided with computing AI
resources by GENCI at IDRIS thanks to the grant R1001016831 (Allocation DALIA)
on the GB200 partition of the Dalia supercomputer. We used a LLM (GLM 5.2) for code generation assistance. \textbf{Compliance with ethical
standards:} The authors have no relevant financial or nonfinancial interests to
disclose. We use the public ICASSP 2024 RF Challenge
datasets~\cite{lancho2024rfchallenge} and the DroneDetect
dataset~\cite{swinney2021dronedetect}.

\bibliographystyle{IEEEbib}
\bibliography{refs}
\end{document}